\documentclass{article} 
\usepackage{iclr2016_conference,times}
\usepackage{hyperref}
\usepackage{url}
\usepackage{amssymb,amsmath,graphicx,amsfonts,bbm,subfigure}
\usepackage{mdwmath}
\usepackage{mdwtab}
\usepackage{algpseudocode}
\usepackage{pifont}
\usepackage{dsfont}

\usepackage{tabularx}
\usepackage{mdwlist}
\usepackage{multirow}
\usepackage{sidecap}

\title{Multimodal Sparse Representation Learning and Applications}

\author{Miriam Cha \& Youngjune L.~Gwon \& H.~T.~Kung\\
John A. Paulson School of Engineering and Applied Sciences\\
Harvard University, Cambridge, MA 02138\\
}

\newcommand{\comment}[1]{}

\DeclareMathAlphabet{\mathitbf}{OML}{cmm}{b}{it}

\newcommand{\eg}{\emph{e.g.}}

\iclrfinalcopy 

\begin{document}

\maketitle

\begin{abstract}
Unsupervised methods have proven effective for discriminative tasks in a single-modality scenario. In this paper, we present a multimodal framework for learning \emph{sparse} representations that can capture semantic correlation between modalities. The framework can model relationships at a higher level by forcing the shared sparse representation. In particular, we propose the use of joint dictionary learning technique for sparse coding and formulate the joint representation for concision, cross-modal representations (in case of a missing modality), and union of the cross-modal representations. Given the accelerated growth of multimodal data posted on the Web such as YouTube, Wikipedia, and Twitter, learning good multimodal features is becoming increasingly important. We show that the shared representations enabled by our framework substantially improve the classification performance under both unimodal and multimodal settings. We further show how deep architectures built on the proposed framework are effective for the case of highly nonlinear correlations between modalities. The effectiveness of our approach is demonstrated experimentally in image denoising, multimedia event detection and retrieval on the TRECVID dataset (audio-video), category classification on the Wikipedia dataset (image-text), and sentiment classification on PhotoTweet (image-text).
\end{abstract}

\section{Introduction}
Human perception works by integrating multiple sensory inputs. Processing different sensory modalities (\eg, vision, auditory,  olfaction) and correlating them improve our perceptual abilities in numerous ways \citep{stein2009}. When different phenomena cause an ambiguity by activating similar features in one modality, features from other modalities can be examined. If one modality is impaired or becomes corrupted, other modalities can help fill in the missing information for robustness. Finally, consensus among modalities can be taken as a reinforcing factor.

Multiple modalities are believed to benefit discriminative machine learning tasks. Using different sensors simultaneously, a scene from the same event can be described in multiple data modalities. For example, consider a multimedia event detection (MED) problem with class names such as ``Dog show," ``Firework," ``Playing fetch with dogs," and ``Shooting a gun." Judging based on the video modality only, ``Dog show" and ``Playing fetch with dogs" come close, featuring both people and dogs in some coordinated actions. However, the two classes are easier to discriminate by incorporating the audio modality in which ``Dog show" is characterized by crowd noise and microphone announcement absent in ``Playing fetch with dog." On the other hand, ``Firework" and ``Shooting a gun" are hard to discriminate with audio, but their visual differences are useful. 

In multimodal feature learning, one wishes to learn good shared representations across heterogeneous data modalities. We can form a union of unimodal features after learning features for each modality separately. This approach, however, has a drawback for being unable to learn patterns that occur jointly or selectively across modalities since unimodal learning emphasizes on relating information within one modality. 

In this paper, we present a sparse coding framework that can model the correlation between modalities. The framework aims to learn relationships at higher level by forcing to share sparse representation. Sparse coding has been recognized widely in machine learning applications such as classification, denoising, and recognition \citep{wright_3}. In particular, it is known that multimedia data can be well-represented as a sparse linear combination of basis vectors. For example, the abundance of unlabeled, photos in the web makes good large dictionaries readily available for sparse coding. 

Similar to other learning methods, sparse coding has been primarily applied under unimodal settings. However, we recognize numerous multimodal approaches for sparse coding from recent literature. These approaches in common aim to learn shared sparse representation for different modalities. \cite{jia_1} exploit structured sparsity to learn a shared latent space of multi-view data (\eg, 2D image + depth). \cite{monaci_1} propose a sparse coding-based scheme that learns bimodal structure in audio-visual speech data. Additionally, \cite{zhuang_1} describe a supervised sparse coding scheme for cross-modal retrieval (\eg, text retrieval from an image query). Our contributions are two-fold. First, we set up an experimental deep architecture built on multiple layers of sparse coding and pooling units. From this, we report promising results on classification with multimodal datasets. Secondly, we demonstrate the performance of multimodal sparse coding in a comprehensive set of applications. In particular, we include our result with TRECVID MED tasks for detecting high-level complex events in user-generated videos. We examine various settings of multimodal sparse coding (detailed in Section 3) using several multimodal datasets of semantically correlated pairs (audio/video and web images/text). Such semantic correlation reveals shared statistical association between modalities and thus can provide complementary information for each other. 

There are existing multimodal learning schemes that are not sparse coding-based. These include audio-visual speech recognition \citep{gurban_2009,papandreou_2007,lucey_2006,ngiam_2011}, sentiment recognition \citep{morency_2011,baecchi,borth_2013}, and image-text retrieval \citep{sohn_2014,feng_2014}. \cite{ngiam_2011} have applied deep stacked autoencoder to learn representations for speech audio coupled with video of the lips. \cite{poirson_2013} have used denoising autoencoder for Flickr photos with associated text. 

In the following sections, we cover the basic principle of sparse coding and extend it to build our multimodal framework on both shallow and deep architectures. We will demonstrate the effectiveness of our approach experimentally with multimedia applications that include image denoising, categorical classification with images and text from Wikipedia, sentiment classification using PhotoTweet, and TRECVID MED.

\section{Preliminaries}
Originated to explain neuronal activations that encode sensory information \citep{sc_1}, sparse coding is an unsupervised method to learn an efficient representation of data using a small number of basis vectors. It has been used to discover higher-level features for data from unlabeled examples. Given a data input $\mathbf{x} \in \mathbb{R}^N$, sparse coding solves for a representation $\mathbf{y} \in \mathbb{R}^K$ while simultaneously updating the dictionary $\mathbf{D} \in \mathbb{R}^{N\times K}$ of $K$ basis vectors as
\begin{align} \label{eq:sc}
\min_{\mathbf{D},\mathbf{y}}\Arrowvert \mathbf{x}-\mathbf{D}\mathbf{y} \Arrowvert_{2}^{2} + \lambda \Arrowvert \mathbf{y} \Arrowvert_1 \quad \mbox{s.t.}~\Arrowvert \mathbf{d}_i \Arrowvert_2 \le 1, \forall i
\end{align} 
where $\mathbf{d}_i$ is $i$th dictionary atom in $\mathbf{D}$, and $\lambda$ is a regularization parameter that penalizes over the $\ell_1$-norm, which induces a sparse solution. With $K > N$, sparse coding typically trains an overcomplete dictionary. This makes the sparse code $\mathbf{y}$ higher in dimension than $\mathbf{x}$, but only $S \ll N$ elements in $\mathbf{y}$ are nonzero. 

Sparse coding can alternatively regularize on the $\ell_0$ pseudo-norm. Finding the sparsest $\ell_0$ solution in general, however, is known to be intractable. Although greedy-$\ell_0$ methods such as orthogonal matching pursuit (OMP) can be used, we only consider Equation~(\ref{eq:sc}) as our choice for sparse coding throughout this paper. We use least angle regression (LARS) and the dictionary learning algorithm by \cite{mairal} from SPAMS toolbox \citep{spams}.

\begin{figure}
\centering
\subfigure[Sparse coding of $a$]{
\includegraphics[width=0.19\textwidth]{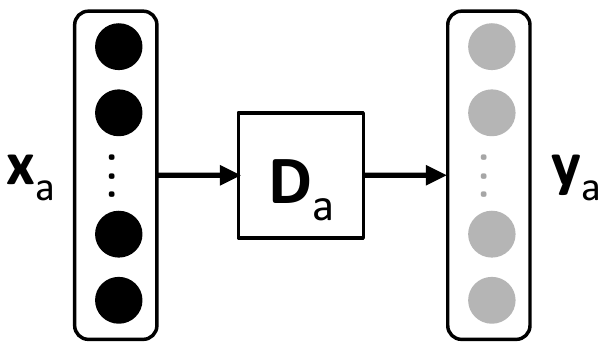}
\label{subfig:uni-a}}
\hspace{20pt}
\subfigure[Sparse coding of $b$]{
\includegraphics[width=0.19\textwidth]{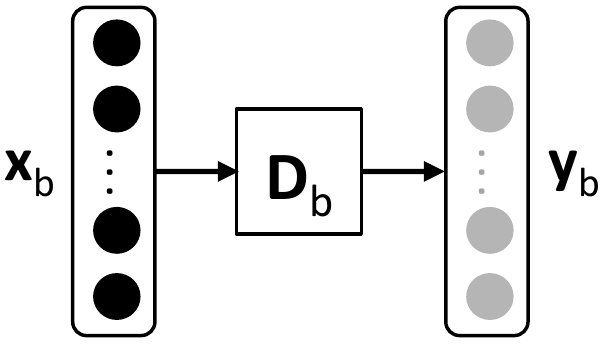}
\label{subfig:uni-b}}
\hspace{20pt}
\subfigure[Feature union]{
\includegraphics[width=0.19\textwidth]{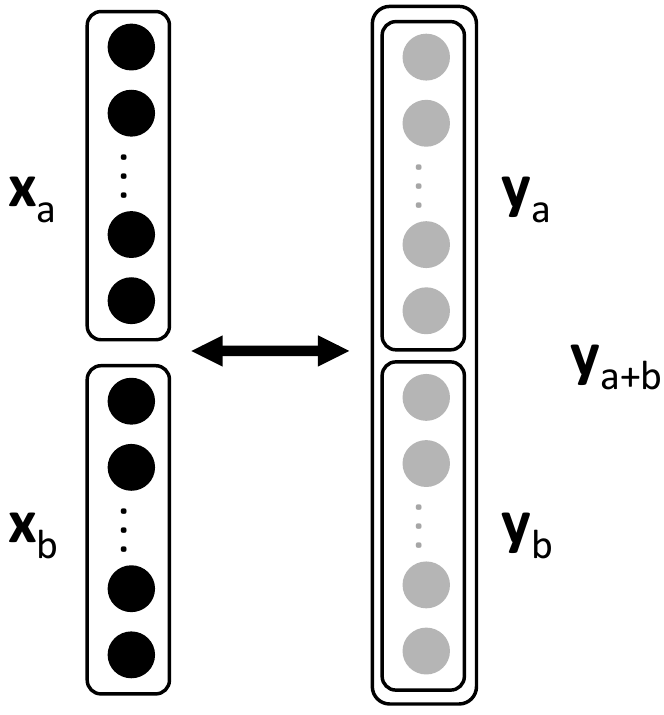}
\label{subfig:uni-fusion}}
\caption{Unimodal sparse coding and feature union}
\label{fig:uni}
\end{figure}

\begin{figure}
\centering
\subfigure[{\scriptsize Joint sparse coding}]{
\includegraphics[width=0.19\textwidth]{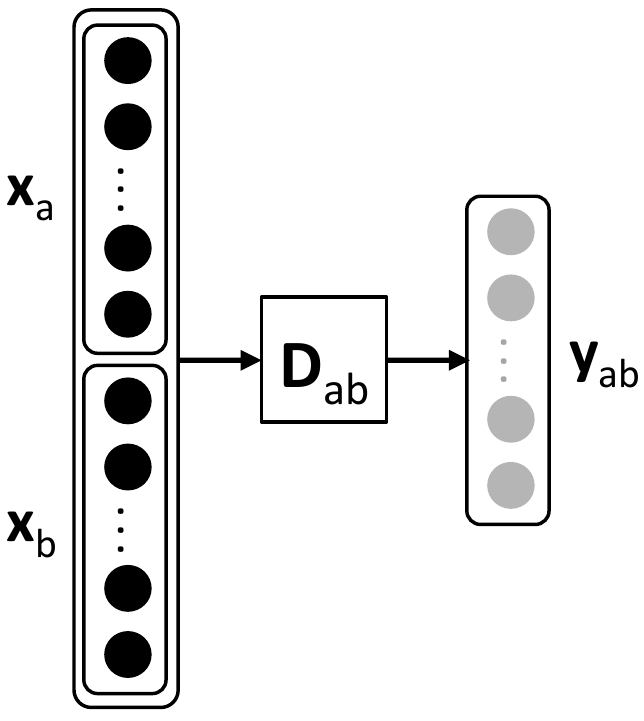}
\label{subfig:multi-ab}}
\hspace{16pt}
\subfigure[Cross-modal by $a$]{
\includegraphics[width=0.19\textwidth]{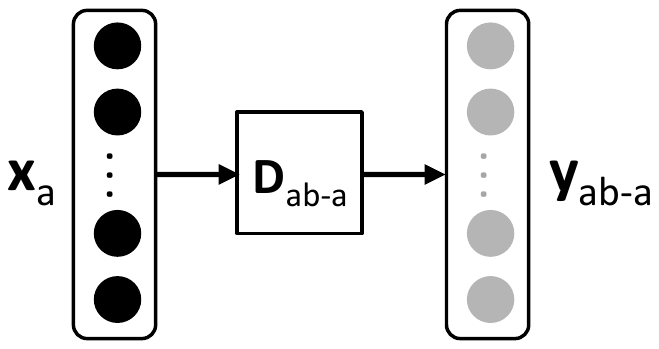}
\label{subfig:multi-a}}
\hspace{16pt}
\subfigure[Cross-modal by $b$]{
\includegraphics[width=0.19\textwidth]{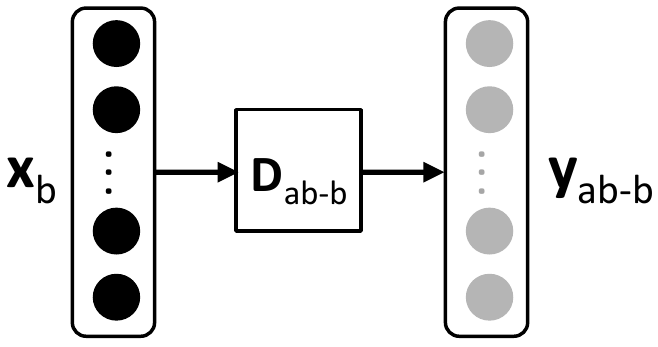}
\label{subfig:multi-b}}
\hspace{16pt}
\subfigure[Feature union]{
\includegraphics[width=0.19\textwidth]{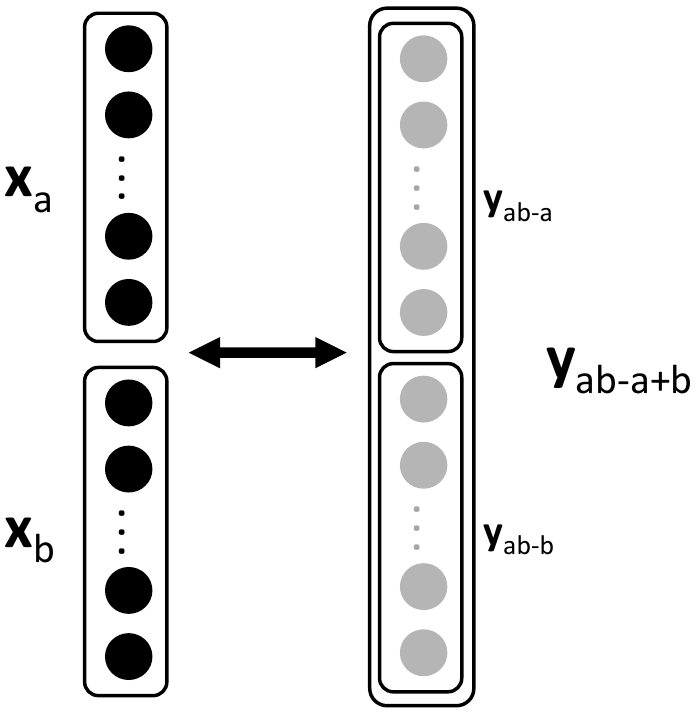}
\label{subfig:multi-fusion}}
\caption{Multimodal sparse coding and feature union}
\label{fig:multi}
\end{figure}

\section{Multimodal Feature Learning via Sparse Coding}
\label{sec:mult_sec}
This section describes our multimodal feature learning schemes for sparse coding. Our schemes are general and can readily be extended for more than two modalities. For clarity of explanation, we use two modalities $a$ and $b$ throughout the section.

\subsection{Parallel unimodal sparse coding}
A straightforward approach for sparse coding with two heterogeneous modalities (\eg, text and images) is to learn a \emph{separate} dictionary of basis vectors for each modality. Figure~\ref{fig:uni} depicts unimodal sparse coding schemes for modalities $a$ and $b$. We learn the two dictionaries $\mathbf{D}_\mathrm{a}$ and $\mathbf{D}_\mathrm{b}$ in parallel \begin{align}
\min_{\mathbf{D}_\mathrm{a},\mathbf{y}_\mathrm{a}^{(i)}} \sum_{i=1}^{n_\mathrm{a}} \Arrowvert \mathbf{x}_\mathrm{a}^{(i)} - \mathbf{D}_\mathrm{a} \mathbf{y}_\mathrm{a}^{(i)} \Arrowvert^2_2 + \lambda \Arrowvert \mathbf{y}_\mathrm{a}^{(i)} \Arrowvert_1, \\
\min_{\mathbf{D}_\mathrm{b},\mathbf{y}_\mathrm{b}^{(i)}} \sum_{i=1}^{n_\mathrm{b}} \Arrowvert \mathbf{x}_\mathrm{b}^{(i)} - \mathbf{D}_\mathrm{b} \mathbf{y}_\mathrm{b}^{(i)} \Arrowvert^2_2 + \lambda \Arrowvert \mathbf{y}_\mathrm{b}^{(i)} \Arrowvert_1.
\end{align} Unimodal sparse coding of $a$ takes in $n_\mathrm{a}$ unlabeled examples $\mathbf{x}_\mathrm{a}^{(i)}$ to train $\mathbf{D}_\mathrm{a}$ while simultaneously computing corresponding sparse code $\mathbf{y}_\mathrm{a}^{(i)}$ under the regularization parameter $\lambda$. (We denote $\mathbf{x}^{(i)}$ the $i$th training example.) Similarly for modality $b$, we train $\mathbf{D}_\mathrm{b}$ from $n_\mathrm{b}$ unlabeled examples $\mathbf{x}_\mathrm{b}^{(i)}$ by computing $\mathbf{y}_\mathrm{b}^{(i)}$. As explained in Figure~\ref{subfig:uni-fusion}, we can form $\mathbf{y}_\mathrm{a+b}=\begin{bmatrix}
\mathbf{y}_\mathrm{a}\\ 
\mathbf{y}_\mathrm{b}
\end{bmatrix}$, a union of the unimodal feature vectors. 



\subsection{Joint multimodal sparse coding}
Union of the unimodal sparse codes $\mathbf{y}_\mathrm{a+b}$ is a simple way to encapsulate the features from both modalities. However, the unimodal training model is flawed since it cannot capture correlations between the two modalities that could be beneficial for inference tasks. To remedy the lack of joint learning, we propose a multimodal sparse coding scheme illustrated in Figure~\ref{subfig:multi-ab}. We use the joint sparse coding technique used in image super-resolution \citep{wright2} \begin{align} \label{eq:jointsc}
\min_{\mathbf{D}_\mathrm{ab},\mathbf{y}_\mathrm{ab}^{(i)}} \sum_{i=1}^{n} \Arrowvert \mathbf{x}_\mathrm{ab}^{(i)} - \mathbf{D}_\mathrm{ab} \mathbf{y}_\mathrm{ab}^{(i)} \Arrowvert^2_2 + \lambda' \Arrowvert \mathbf{y}_\mathrm{ab}^{(i)} \Arrowvert_1.
\end{align} 
Here, we train with concatenated input vectors 
$\mathbf{x}_\mathrm{ab}^{(i)} = \begin{bmatrix}
\frac{1}{\sqrt{N_\mathrm{a}}}\mathbf{x}_\mathrm{a}^{(i)}\\ 
\frac{1}{\sqrt{N_\mathrm{b}}}\mathbf{x}_\mathrm{b}^{(i)}
\end{bmatrix}$, where $N_\mathrm{a}$ and $N_\mathrm{b}$ are dimensions of $\mathbf{x}_\mathrm{a}$ and $\mathbf{x}_\mathrm{b}$, respectively. As an interesting property, we can decompose the multimodal dictionary as $\mathbf{D}_\mathrm{ab} = \begin{bmatrix}
\frac{1}{\sqrt{N_\mathrm{a}}}\mathbf{D}_\mathrm{ab\mbox{-}a}\\ 
\frac{1}{\sqrt{N_\mathrm{b}}}\mathbf{D}_\mathrm{ab\mbox{-}b}
\end{bmatrix}$ to perform the cross-modal sparse coding \begin{align} \label{eq:ja}
\min_{\mathbf{y}_\mathrm{ab\mbox{-}a}^{(i)}} \sum_{i=1}^{n_\mathrm{a}} \Arrowvert \mathbf{x}_\mathrm{a}^{(i)} - \mathbf{D}_\mathrm{ab\mbox{-}a} \mathbf{y}_\mathrm{ab\mbox{-}a}^{(i)} \Arrowvert^2_2 + \lambda'' \Arrowvert \mathbf{y}_\mathrm{ab\mbox{-}a}^{(i)} \Arrowvert_1,\\ \label{eq:jb} \min_{\mathbf{y}_\mathrm{ab\mbox{-}b}^{(i)}} \sum_{i=1}^{n_\mathrm{b}} \Arrowvert \mathbf{x}_\mathrm{b}^{(i)} - \mathbf{D}_\mathrm{ab\mbox{-}b} \mathbf{y}_\mathrm{ab\mbox{-}b}^{(i)} \Arrowvert^2_2 + \lambda'' \Arrowvert \mathbf{y}_\mathrm{ab\mbox{-}b}^{(i)} \Arrowvert_1.
\end{align} In principle, joint sparse coding in Equation~(\ref{eq:jointsc}) combines the objectives of Equations~(\ref{eq:ja}) and (\ref{eq:jb}), forcing the sparse codes $\mathbf{y}_\mathrm{ab\mbox{-}a}^{(i)}$ and $\mathbf{y}_\mathrm{ab\mbox{-}b}^{(i)}$ to share the same values when $\lambda' = (\frac{1}{N_\mathrm{a}} + \frac{1}{N_\mathrm{b}})\lambda''$. Ideally, we could have $\mathbf{y}_\mathrm{ab}^{(i)} = \mathbf{y}_\mathrm{ab\mbox{-}a}^{(i)} = \mathbf{y}_\mathrm{ab\mbox{-}b}^{(i)}$, although empirical values determined by the three different optimizations would differ in reality. According to \cite{wright2}, $\mathbf{x}_\mathrm{a}$ and $\mathbf{x}_\mathrm{b}$ are highly correlated as the low and high resolution images are originated from the same source. However, if $\mathbf{x}_\mathrm{a}$ and $\mathbf{x}_\mathrm{b}$ come from two different modalities, their correlation is present in semantics. Thus, equality assumption among $\mathbf{y}_\mathrm{ab}^{(i)}$, $\mathbf{y}_\mathrm{ab\mbox{-}a}^{(i)}$, and $\mathbf{y}_\mathrm{ab\mbox{-}b}^{(i)}$ is even less likely to be met. For that reason, we introduce cross-modal sparse coding that captures weak correlation between heterogeneous modalities, resulting sparse code that is more discriminative. Cross-modal sparse coding first trains a joint dictionary $\mathbf{D}_\mathrm{ab}$. Then in test time, cross-modal sparse codes are computed using sub-dictionaries $\mathbf{D}_\mathrm{ab\mbox{-}a}$ and $\mathbf{D}_\mathrm{ab\mbox{-}b}$. Various feature formation possibilities on multimodal sparse coding---joint, cross-modal, and union of cross-modal sparse codes are explained in Figure~\ref{fig:multi}. 

\subsection{Deep multimodal sparse coding}
So far, we have only considered shallow learning architectures using a single layer of sparse coding and dictionary learning. The shallow architecture is capable of learning the features jointly, but it may not be sufficient to capture the complex semantic correlations between the modalities fully. We expect modalities with high semantic correlation to be stable in hierarchical architecture as it can utilize a large number of hidden layers and parameters to extract more meaningful high-level representation from modalities. Composing higher-level representation using low-level features should be advantageous to contextual data such as human language, speech, audio, and sequence of image patterns. Hierarchical composition of sparse codes has shown to help unveil separability of data invisible from their lower-level features in unimodal settings \citep{bo_2013,yu_2011}. Therefore, we consider deep architectures for multimodal sparse coding. 

\begin{figure}
\centering
\subfigure[Deep unimodal sparse coding before joint sparse coding]{%
\includegraphics[width=0.30\textwidth]{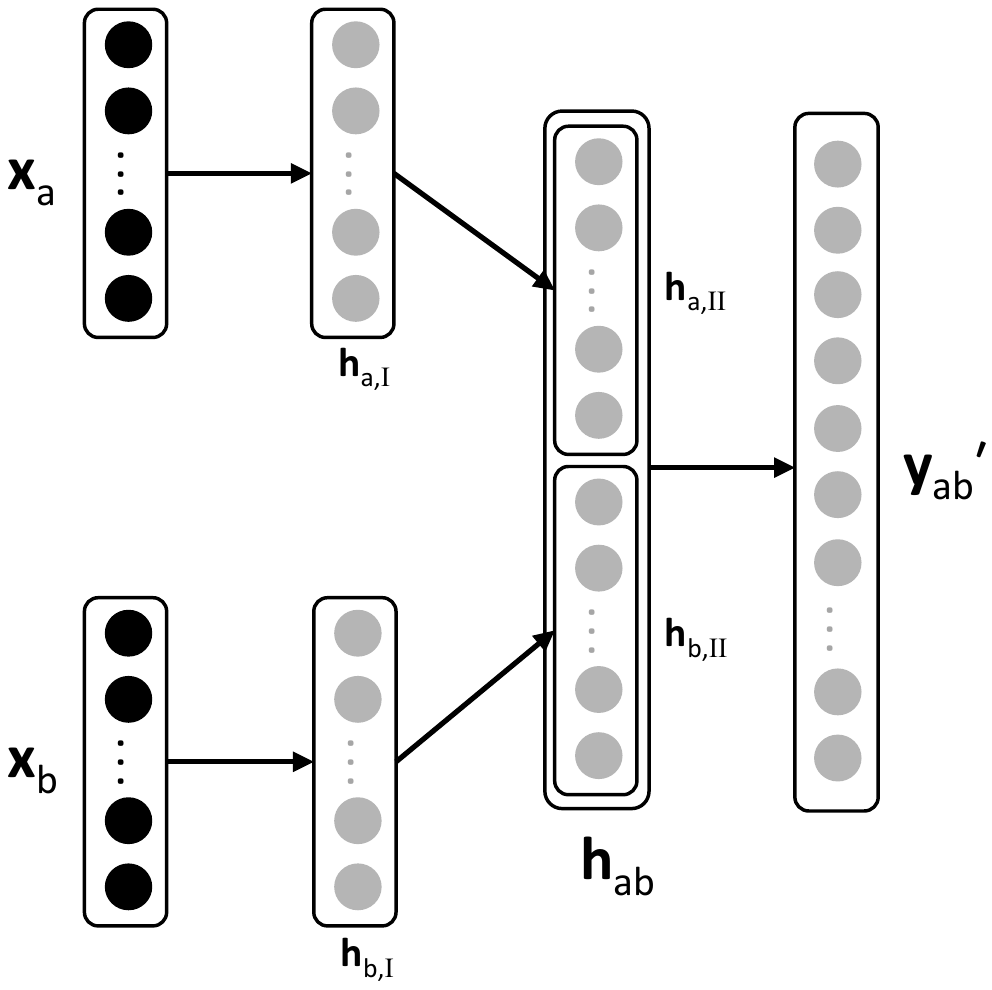}
\label{subfig:deep1}}
\hspace{16pt}
\subfigure[Deep unimodal and deep joint sparse coding]{%
\includegraphics[width=0.30\textwidth]{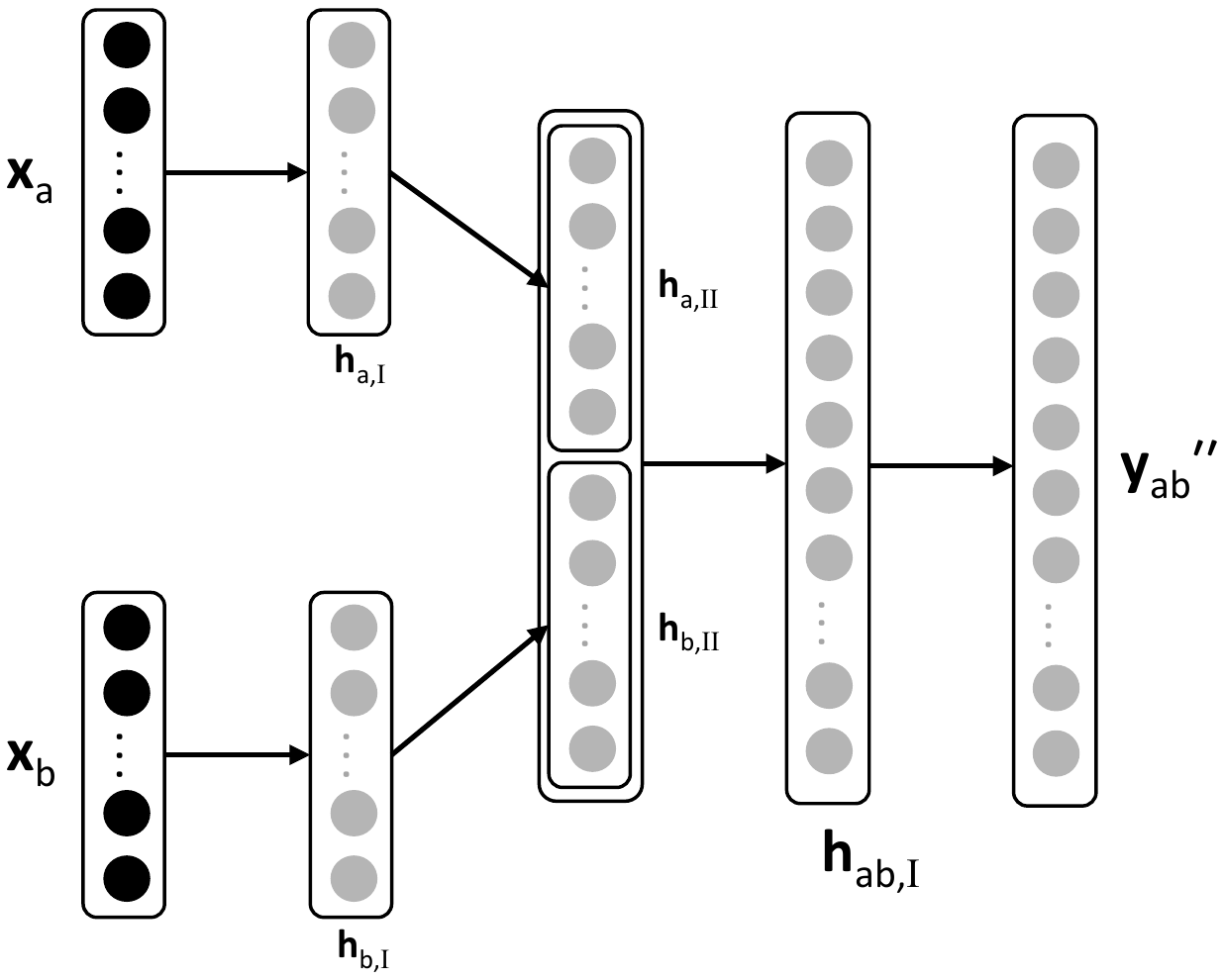}
\label{subfig:deep2}}
\caption{Deep learning approaches for multimodal sparse coding}
\label{fig:deep}
\end{figure}

In Figure~\ref{fig:deep}, we propose two possible architectures. \cite{ngiam_2011} report their RBM-based approaches beneficial when applying deep learning on each modality before joint training. Adopting their configuration, we use (at least) two layers of sparse coding for each modality followed by the joint sparse coding as illustrated in Figure~\ref{subfig:deep1}. We write two-layer sparse coding for modality $a$ \begin{align} \min_{\mathbf{D}_\mathrm{a,I},\mathbf{y}_\mathrm{a,I}^{(i)}} \sum_{i=1}^{n_\mathrm{a}} \Arrowvert \mathbf{x}_\mathrm{a}^{(i)} - \mathbf{D}_\mathrm{a,I} \mathbf{y}_\mathrm{a,I}^{(i)} \Arrowvert^2_2 + \gamma \Arrowvert \mathbf{y}_\mathrm{a,I}^{(i)} \Arrowvert_1, \\
\{\mathbf{y}_\mathrm{a,I}^{(i)},\dots,\mathbf{y}_\mathrm{a,I}^{(i+M_\mathrm{I}-1)}\} \overset{_\mathrm{pool}}{\longrightarrow} \mathbf{h}_\mathrm{a,I}^{(j)},\\
\min_{\mathbf{D}_\mathrm{a,II},\mathbf{y}_\mathrm{a,II}^{(j)}} \sum_{j=1}^{n_\mathrm{a}'} \Arrowvert \mathbf{h}_\mathrm{a}^{(j)} - \mathbf{D}_\mathrm{a,II} \mathbf{y}_\mathrm{a,II}^{(j)} \Arrowvert^2_2 + \gamma \Arrowvert \mathbf{y}_\mathrm{a,II}^{(j)} \Arrowvert_1, \\
\{\mathbf{y}_\mathrm{a,II}^{(j)},\dots,\mathbf{y}_\mathrm{a,I}^{(j+M_\mathrm{II}-1)}\} \overset{_\mathrm{pool}}{\longrightarrow} \mathbf{h}_\mathrm{a,II}^{(k)}.
\end{align} We denote $\mathbf{D}_\mathrm{a,I}$ and $\mathbf{D}_\mathrm{a,II}$ the dictionaries learned by the two sparse-coding layers for $a$, \emph{unpooled} sparse codes $\mathbf{y}_\mathrm{a,I}$ and $\mathbf{y}_\mathrm{a,II}$, and the hidden activations $\mathbf{h}_\mathrm{a,I}$ and $\mathbf{h}_\mathrm{a,II}$ by max (or average) pooling sparse codes. Max pooling factors $M_\mathrm{I}$ and $M_\mathrm{II}$ refer to the number of sparse codes aggregated to one pooled representation. They are determined empirically. Since sparse coding takes in dense input vectors and produces sparse output vectors, it is a poor fit for multilayering. Hence, we interlace nonlinear pooling units between sparse coding layers and aggregate multiple sparse vectors nonlinearly to a dense vector before passing onto next layer. Similar to modality $a$, we can work out $\mathbf{D}_\mathrm{b,I}$, $\mathbf{D}_\mathrm{b,II}$, $\mathbf{y}_\mathrm{b,I}$, $\mathbf{y}_\mathrm{b,II}$, $\mathbf{h}_\mathrm{b,I}$, and $\mathbf{h}_\mathrm{b,II}$ for modality $b$. Ultimately, the joint feature representation $\mathbf{y}_\mathrm{ab}'$ in Figure~\ref{subfig:deep1} is learned by \begin{align}
\min_{\mathbf{D}_\mathrm{ab}',\mathbf{y}_\mathrm{ab}'^{(k)}} \sum_{k=1}^{n'} \Arrowvert \mathbf{h}_\mathrm{ab}^{(k)} - \mathbf{D}_\mathrm{ab}' \mathbf{y}_\mathrm{ab}'^{(k)} \Arrowvert^2_2 + \gamma' \Arrowvert \mathbf{y}_\mathrm{ab}'^{(k)} \Arrowvert_1
\end{align} where we input the concatenated deep hidden representations $\mathbf{h}_\mathrm{ab}^{(k)} = \begin{bmatrix}
\mathbf{h}_\mathrm{a,II}^{(k)}\\ 
\mathbf{h}_\mathrm{b,II}^{(k)}
\end{bmatrix}$ to train $\mathbf{D}_\mathrm{ab}'$.

Moreover, we consider additional joint sparse coding layer as illustrated in Figure~\ref{subfig:deep2}. Again, we form dense hidden activations $\mathbf{h}_\mathrm{ab,I}$ by pooling multiple $\mathbf{y}_\mathrm{ab}'$ vectors. We compute the deeper representation $\mathbf{y}_\mathrm{ab}''$ by \begin{align}
\min_{\mathbf{D}_\mathrm{ab}'',\mathbf{y}_\mathrm{ab}''^{(l)}} \sum_{l=1}^{n''} \Arrowvert \mathbf{h}_\mathrm{ab,I}^{(l)} - \mathbf{D}_\mathrm{ab}'' \mathbf{y}_\mathrm{ab}''^{(l)} \Arrowvert^2_2 + \gamma'' \Arrowvert \mathbf{y}_\mathrm{ab}''^{(l)} \Arrowvert_1.
\end{align}

\section{Experiments}
\subsection{Image denoising}
In our first experiment, we evaluate the image denoising problem, where zero-mean Gaussian noise is to be removed from a given image. The proposed learning algorithm is used to jointly learn associations between clean and noisy images. We consider clean and its noisy counterpart as two input modalities and recover a clean image from a noisy one. We randomly select 2,500 images from CIFAR-10 \citep{krizhevsky_2009} and add zero-mean Gaussian noise with a range of standard deviation $\sigma$ to generate noisy images. We use 2,000 pairs of clean and noisy images for training joint dictionary and 500 pairs for testing. To test, we input a noisy image and compute cross-modal (noisy to clean) sparse code to recover a clean estimate.  We compare the performance of the joint multimodal sparse coding with denoising autoencoder (DAE) \citep{vincent_2008}. Table~\ref{tab:psnr_tab} summarizes denoising results for the joint sparse coding and DAE. In this experiment, the dictionaries used were of size $N \times K$, designed to handle clean and noisy image patches of size 4 $\times$ 4 pixels ($N=32$, $K = 300$). Every result reported is an average over 5 experiments. For different realization of the noise, joint sparse coding improves the quality of noisy images, and DAE often achieves comparable results. 

\begin{SCfigure}
  \centering
  \caption{Example of the denoising results with $\sigma=0.01$. From top to bottom, we visualize clean original image, noisy image, denoised images using DAE and joint sparse coding.}
  \includegraphics[width=0.47\textwidth]%
    {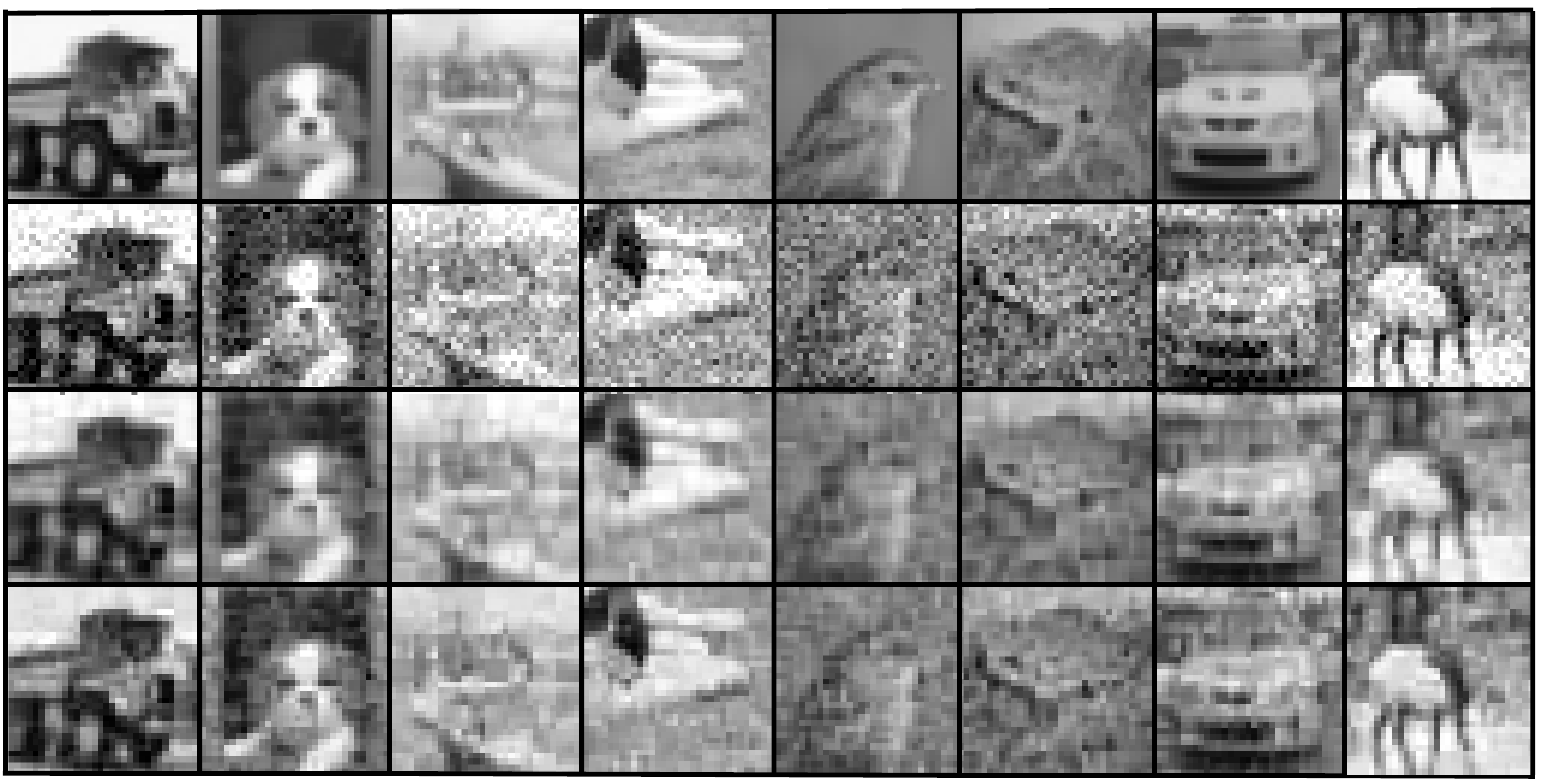}
    \label{fig:denoise} 
\end{SCfigure}

In Figure~\ref{fig:denoise}, we visualize the generated clean images from noisy images with $\sigma=0.01$. Notice that joint sparse coding can learn shared association between clean and noisy images as in DAE to achieve comparable results. 

\begin{table}[t]
\footnotesize
\centering
\caption{Peak signal-to-noise ratio results (in decibels) on image denoising using denoising autoencoder and joint sparse coding with a range of standard deviation $\sigma$.}
\label{tab:psnr_tab}
\begin{tabular}{c|c|c}
\hline
$\sigma$/PSNR & DAE & Joint sparse coding \\ \hline
0.001/30.08 & 28.87 & 31.08 \\
0.005/23.20 & 25.61 & 25.71 \\
0.01/20.28 & 24.14 & 23.87 \\
0.1/11.52 & 18.74 & 18.19 \\ \hline
\end{tabular}
\end{table}

\subsection{Audio and video}
In this section, we apply our multimodal sparse coding schemes to multimedia event detection (MED) tasks. MED aims to identify complex activities encompassing various human actions, objects, and their interactions at different places and time. MED is considered more difficult than concept analysis such as action recognition and has received significant attention in computer vision and machine learning research.

\textbf{Dataset, tasks, and metrics.} We use the TRECVID 2014 dataset \citep{trecvidmed14} to evaluate our schemes. We consider the event detection and retrieval tasks using the 10Ex and 100Ex data scenarios, where 10Ex gives 10 multimedia examples per event, and 100 examples per event for 100Ex. There are 20 event classes (E021 to E040) with event names such as ``Bike trick," ``Dog show," and ``Marriage proposal." We evaluate the MED peformance trained on the features learned from our unimodal and multimodal sparse coding schemes. In particular, we compute classification accuracy and mean average precision (mAP) metrics according to the NIST standard on the following experiments: 1) cross-validation on 10Ex; and 2) train using 10Ex and test with 100Ex.

\begin{figure}
\centering
\subfigure[Keyframe extraction]{
\includegraphics[width=0.4\textwidth]{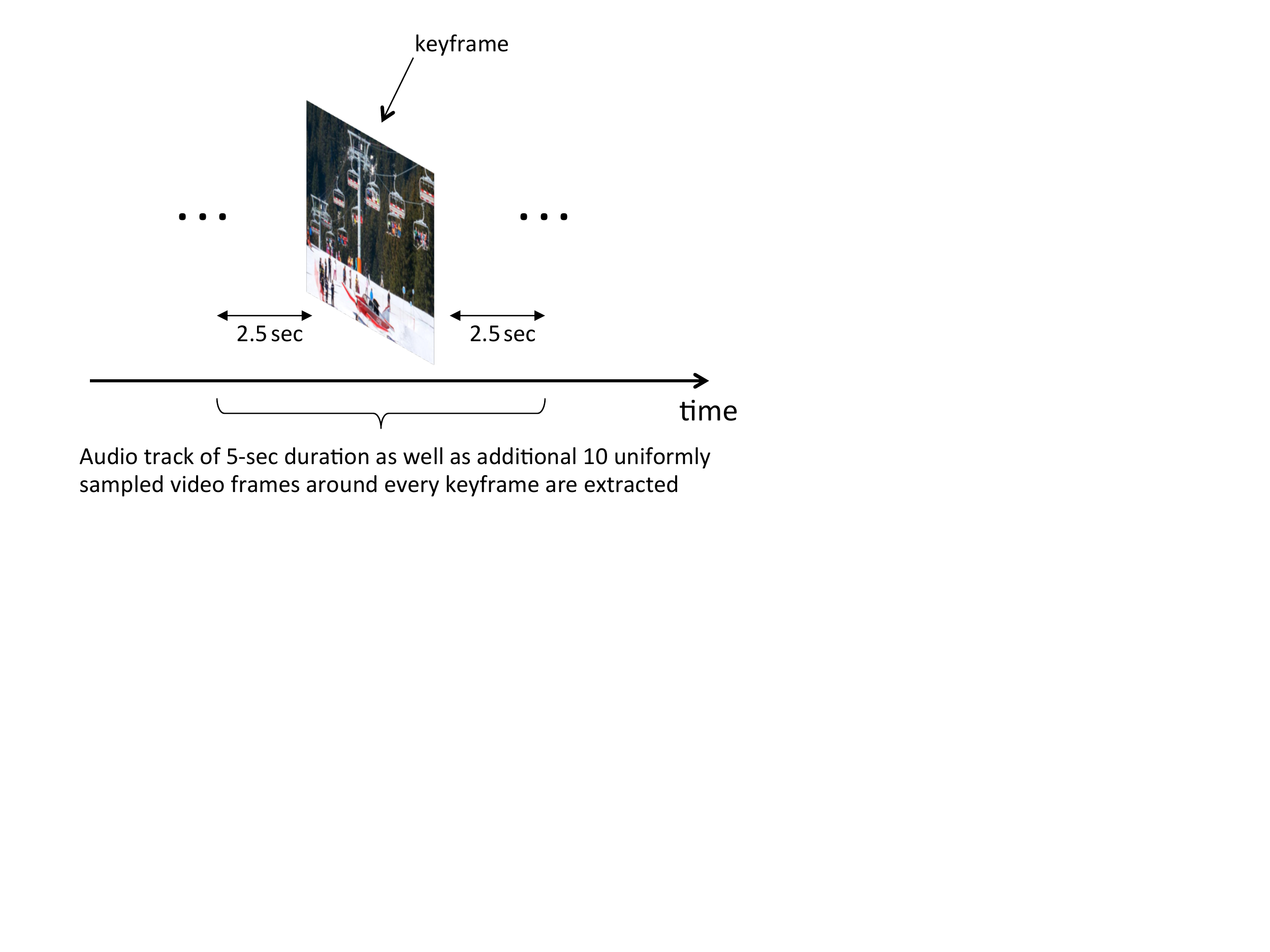}
\label{subfig:keyframe-ext}}
\hspace{30pt}
\subfigure[Audio]{
\includegraphics[height=0.4\textwidth]{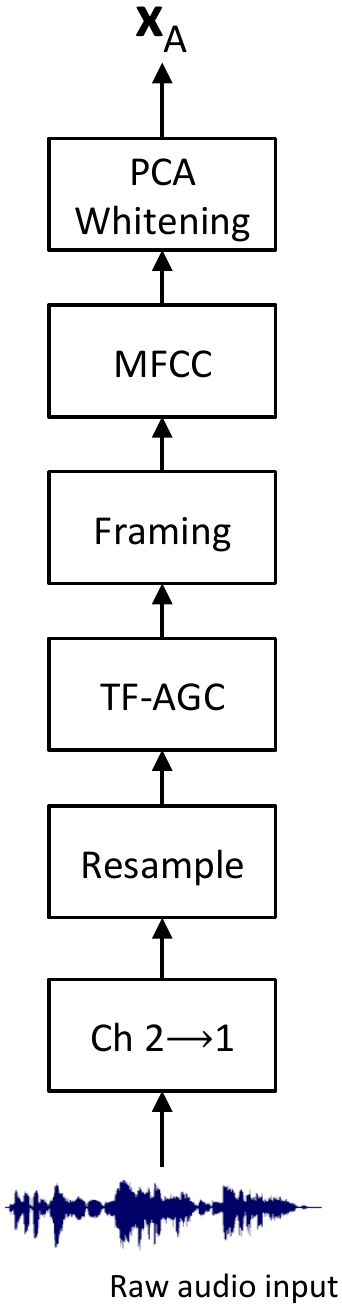}
\label{subfig:audio-preproc}}
\hspace{30pt}
\subfigure[Video]{
\includegraphics[height=0.32\textwidth]{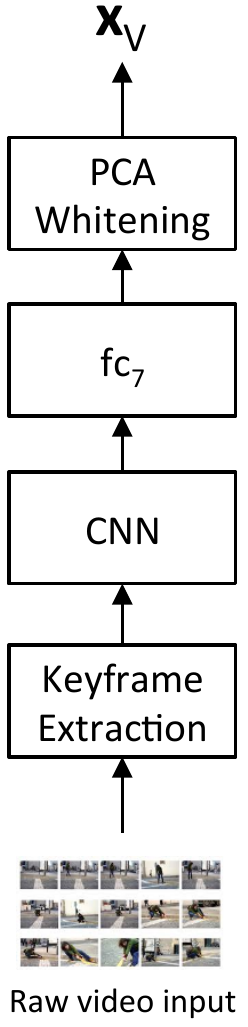}
\label{subfig:video-preproc}}
\caption{TRECVID audio-video data processing}
\label{fig:trec_av}
\end{figure}

\textbf{Data processing.} For processing efficiency and scalability, we use keyframe-based feature extraction for audio-video data. We apply a simple two-pass algorithm that computes color histogram difference of any two successive frames and determines a keyframe candidate based on the threshold calculated on the mean and standard deviation of the histogram differences. We examine the number of different colors present in the keyframe candidates and discard the ones with less than 26 colors (\eg, all-black or all-white) to ensure non-blank keyframes.

Around each keyframe, we extract 5-sec audio data and additional 10 uniformly sampled video frames within the duration as illustrated in Figure~\ref{subfig:keyframe-ext}. If extracted audio is stereo, we take only its left channel. The audio waveform is resampled to 22.05\,kHz and regularized by the time-frequency automatic gain control (TF-AGC) to balance the energy in sub-bands. We form audio frames using a 46-msec Hann window with 50\% overlap between successive frames for smoothing. For each frame, we compute 16 Mel-frequency cepstral coefficients (MFCCs) as the low-level audio feature. In addition, we append 16 delta cepstral and 16 delta-delta cepstral coefficients, which make our low-level audio feature vectors 48 dimensional. We apply PCA whitening before unsupervised learning. The complete audio preprocessing steps are described in Figure~\ref{subfig:audio-preproc}. 

For video preprocessing, we have tried out pretrained convolutional neural network (CNN) models and ended up choosing \texttt{VGG\_ILSVRC\_19\_layers} by University of Oxford's Visual Geometry Group (VGG) \citep{vgg} for the ImageNet Large-scale Visual Recognition Challenge (ILSVRC). As depicted in Figure~\ref{subfig:video-preproc}, we run the CNN feedforward passes with the extracted video frames and take 4,096-dimensional hidden activation from $\mathrm{fc}_7$, the highest hidden layer before the final ReLU. By PCA whitening, we reduce the dimensionality to 128.

\textbf{Feature learning for MED.} We build feature vectors by sparse coding the preprocessed audio and video data. We use the number of basis vectors $K = 512$ same for all dictionaries under unimodal and multimodal sparse coding schemes. We aggregate sparse codes around each keyframe of a training example by max pooling to form feature vectors for classification. (The pooled feature vectors can scale to file level.) We train linear, 1-vs-all SVM classifiers for each event whose hyper-parameters are determined by 5-fold cross-validation on 10Ex.

\textbf{Other feature learning schemes for comparison.} We consider other unsupervised methods to learn audio-video features for comparison. We report the results for Gaussian mixture model (GMM) and restricted Boltzmann machine (RBM) under similar unimodal and multimodal settings. For GMM, we use the expectation-maximization (EM) to fit the preprocessed input vectors audio-only, video-only, concatenated audio-video in 512 mixtures and form GMM supervectors \citep{gmmsv} containing posterior probabilities with respect to each Gaussian. The max-pooled GMM supervectors are used to train SVMs. We adopt the shallow bimodal pretraining model by \cite{ngiam_2011} for RBM. For fairness, we use the hidden layer of a size 512, and the max-pooled activations are used to train SVMs. A target sparsity of 0.1 is applied to both GMM and RBM.

\textbf{Results.} Table~\ref{tab:res-sc} presents the classification accuracy and mAP performance of unimodal and multimodal sparse coding schemes. For the 10Ex/100Ex experiment, we have used the best parameter setting from the 10Ex cross-validation to test 100Ex examples. In general, we observe that the union of unimodal audio and video feature vectors perform better than using only unimodal or cross-modal features. The multimodal union scheme performs better than the joint schemes. The union schemes, however, double feature dimensionality since our union operation concatenates the two feature vectors. Joint feature vector is an economical way of combining both the audio and video features while keeping the same dimension as audio-only or video-only.

In Table~\ref{tab:others}, we report the mean accuracy and mAP for GMM and RBM under the union and joint feature learning schemes on the 10Ex/100Ex experiment. Our results show that sparse coding is better than GMM by 5--6\% in accuracy and 7--8\% in mAP. However, we find that the performance of RBM is on par with sparse coding. This leaves us a good next step to develop joint feature learning scheme for RBM.  

\vspace{-.2in}
\begin{table}[h!]
\footnotesize
\centering
\caption{Mean accuracy and mAP performance of sparse coding schemes}
\begin{tabular}{|l|ccc|cccc|}
\hline
\multirow{3}{*}{} & \multicolumn{3}{c|}{Unimodal} & \multicolumn{4}{c|}{Multimodal}\\ 
& Audio-only & Video-only & Union & Audio & Video & Joint & Union \\ 
& {\scriptsize (Fig.~\ref{subfig:uni-a})} & {\scriptsize (Fig.~\ref{subfig:uni-b})} & {\scriptsize (Fig.~\ref{subfig:uni-fusion})} & {\scriptsize (Fig.~\ref{subfig:multi-a})} & {\scriptsize (Fig.~\ref{subfig:multi-b})} & {\scriptsize (Fig.~\ref{subfig:multi-ab})} & {\scriptsize (Fig.~\ref{subfig:multi-fusion})} \\ \hline \hline
\begin{tabular}[c]{@{}l@{}} Mean accuracy\\ (cross-val. 10Ex)\end{tabular} & 69\% & 86\% & \textbf{89\%} & 75\% & 87\% & 90\% & \textbf{91\%} \\ \hline
\begin{tabular}[c]{@{}l@{}} mAP\\ (cross-val. 10Ex)\end{tabular} & 20.0\% & 28.1\% & \textbf{34.8\%} & 27.4\% & 33.1\% & 35.3\% & \textbf{37.9\%}\\ \hline \hline
\begin{tabular}[c]{@{}l@{}} Mean accuracy\\ (10Ex/100Ex)\end{tabular} & 56\% & 64\% & \textbf{71\%} & 58\% & 67\% & 71\% & \textbf{74\%} \\ \hline
\begin{tabular}[c]{@{}l@{}} mAP\\ (10Ex/100Ex)\end{tabular} & 17.3\% & 28.9\% & \textbf{30.5\%} & 23.6\% & 28.0\% & 28.4\% & \textbf{33.2\%}\\ \hline
\end{tabular}
\label{tab:res-sc}
\end{table}

\begin{table}
\footnotesize
\centering
\caption{Mean accuracy and mAP performance for GMM and RBM on 10Ex/100Ex}
\begin{tabular}{|l|c|c|}
\hline
Feature learning schemes  & Mean accuracy & mAP \\ \hline
\begin{tabular}[c]{@{}l@{}} Union of unimodal GMM\\features (Figure~\ref{subfig:uni-fusion})\end{tabular}  & 66\% & 23.5\% \\ \hline
\begin{tabular}[c]{@{}l@{}} Multimodal joint GMM\\feature (Figure~\ref{subfig:multi-ab})\end{tabular}    & 68\% & 25.2\% \\ \hline \hline
\begin{tabular}[c]{@{}l@{}} Union of unimodal RBM\\features (Figure~\ref{subfig:uni-fusion})\end{tabular}  & 70\% & 30.1\% \\ \hline
\begin{tabular}[c]{@{}l@{}} Multimodal joint RBM\\feature (Figure~\ref{subfig:multi-ab})\end{tabular} & 72\% & 31.3\% \\ \hline
\end{tabular}
\label{tab:others}
\end{table}

\subsection{Images and text}

In the third set of experiments, we consider learning shared associations between images and text for classification. We evaluate our methods on image-text datasets: Wikipedia \citep{rasiwasia_2010} and PhotoTweet \citep{borth_2013}.

\subsubsection{Wikipedia}
Each article in Wikipedia dataset contains paragraphs of text on a subject with a corresponding picture. These text-image pairs are annotated with a label from 10 categories: art, biology, geography, history, literature, media, music, royalty, sport, and warfare. The corpus contains a total of 2,866 documents. We have split the documents into five folds. Along with the dataset, the authors have supplied the features for images and text. Each image is represented by a histogram of a 128-codeword SIFT \citep{lowe_2004} features, and each text is represented by a histogram of a 10-topic latent Dirichlet allocation (LDA) model \citep{blei_2003}. We use the SIFT features for images and LDA features for text as input to sparse coding algorithms, and then train the 1-vs-all SVMs. After that, we predict the labels of the test image-text pairs and report the accuracy. In this experiment, we have used $K \approx 3N$. 

Table~\ref{tab:wiki} reports the results for various feature representations. In comparison between unimodal sparse coding of images or text alone, text features outperform the image features. In Wikipedia, category membership is mostly driven by text. Categorization solely based on image is ambiguous and difficult even for a human. Thus, it is expected that image features would have lower accuracy than text features. The union of  unimodal image and text features (Table ~\ref{tab:wiki}c) further improves the accuracy. Joint sparse coding (Table~\ref{tab:wiki}d) is able to learn multimodal features that go beyond simply concatenating the two unimodal features. 

We notice that learning shared association between image and text can improve classification accuracy even when only a single modality is available for training and testing. Cross-modal by images (Table~\ref{tab:wiki}e) improves accuracy by 4.35\% compared to unimodal sparse coding of images (Table~\ref{tab:wiki}a), and cross-modal by text (Table~\ref{tab:wiki}f) achieves 4.54\% higher accuracy than unimodal sparse coding of text (Table~\ref{tab:wiki}b). When the cross-modal features by images and text are concatenated, it outperforms the other feature combinations. 

For all feature representations, multimodal features significantly outperform the unimodal features. This shows that the learned shared association between multiple modalities are useful for both cases when a single or multiple modalities are available for training and testing.

\begin{figure}[t]
\centering 
\includegraphics[width=.48\textwidth]{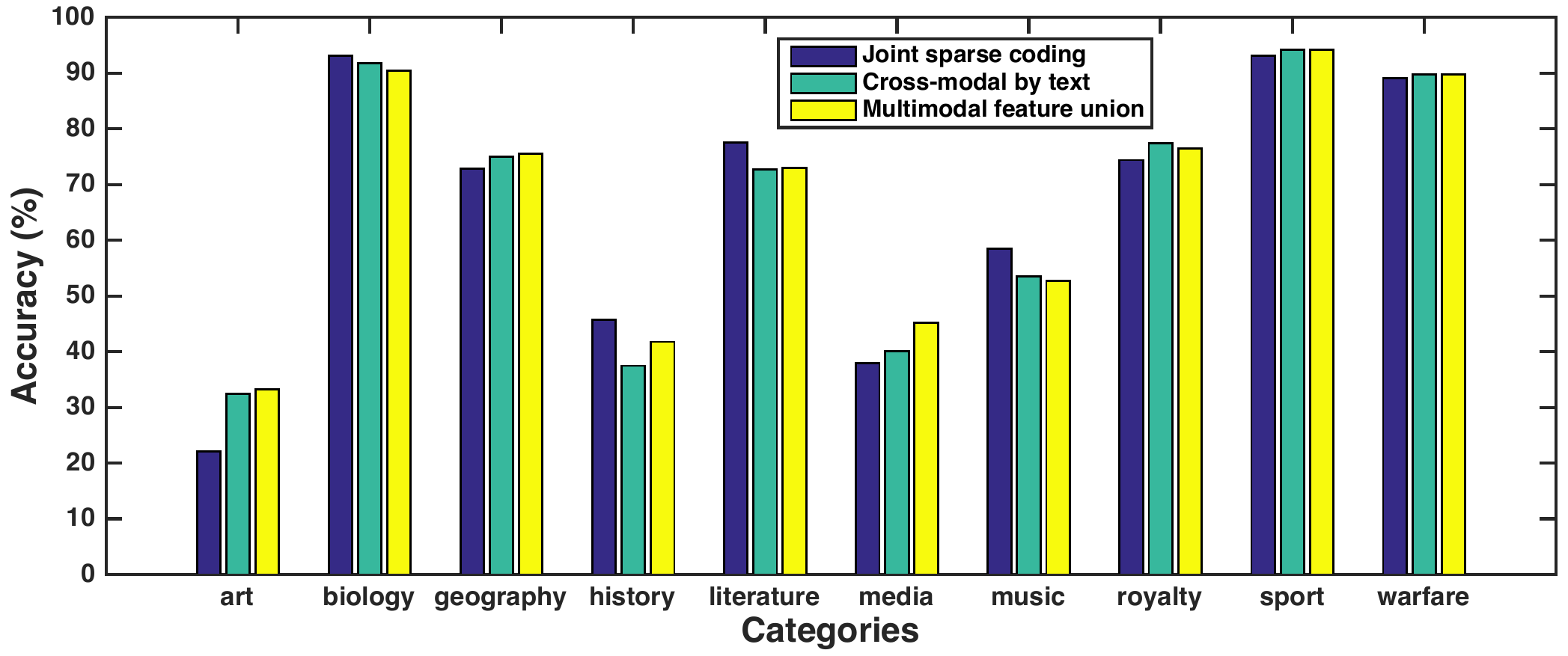} 
\caption{Category classification comparison of joint sparse coding, cross-modal by text, and multimodal feature union on Wikipedia dataset}
\label{fig:wiki_accs} 
\end{figure}

Figure~\ref{fig:wiki_accs} compares classification accuracies of joint sparse coding (Table~\ref{tab:wiki}d), cross-modal by text (Table~\ref{tab:wiki}f), and multimodal feature union (Table~\ref{tab:wiki}g) in 10 categories. Although multimodal feature union does not achieve the best accuracy value for several categories, the classification accuracy is almost close to the best one.

\begin{table}[t]
\footnotesize
\centering
\caption{Classification performance for image-text classification on Wikipedia dataset.}
\label{tab:wiki}
\begin{tabular}{l|c}
\hline
Feature Representation                                                         & Accuracy \\ \hline
\begin{tabular}[c]{@{}l@{}}(a) Sparse coding of images\\ (Figure~\ref{fig:uni}a)\end{tabular}  & 16.93\%      \\
\begin{tabular}[c]{@{}l@{}}(b) Sparse coding of text\\ (Figure~\ref{fig:uni}b)\end{tabular}    & 61.89\%      \\
\begin{tabular}[c]{@{}l@{}}(c) Unimodal feature union\\ (Figure~\ref{fig:uni}c)\end{tabular}   & \textbf{63.38\%}      \\ \hline\hline
\begin{tabular}[c]{@{}l@{}}(d) Joint sparse coding\\ (Figure~\ref{fig:multi}a)\end{tabular}      & 66.44\%      \\
\begin{tabular}[c]{@{}l@{}}(e) Cross-modal by images\\ (Figure~\ref{fig:multi}b)\end{tabular}    & 21.28\%      \\
\begin{tabular}[c]{@{}l@{}}(f) Cross-modal by text\\ (Figure~\ref{fig:multi}c)\end{tabular}      & 66.43\%     \\
\begin{tabular}[c]{@{}l@{}}(g) Multimodal feature union\\ (Figure~\ref{fig:multi}d)\end{tabular} & \textbf{67.23\%}      \\ \hline
\end{tabular}
\end{table}

\subsubsection{PhotoTweet}
PhotoTweet dataset includes 603 Twitter messages (\emph{tweets}) with their associated photos. The benchmark is designed for Twitter user's sentiment prediction using visual and textual features. PhotoTweet is collected in November 2012 via the PeopleBrowsr API. Ground truths labels are obtained by Amazon Mechanic Turk annotation, resulting in 470 positive and 133 negative sentiments. The authors of the dataset have partitioned the dataset into five folds for cross-validation.

We represent the textual data using binary \emph{bag-of-words} embedding, producing 2,688 dimensions vector. We process the embedded tweet data to consecutive patches of a configurable size, $N_\mathrm{a}=42$. The dimension of resulting patches are reduced to 12 dimensions with PCA whitening. For images, we consider a \emph{per-image} feature vector from non-overlapping patches drawn from a receptive field with width $w = 4$ pixels. Thus, each (colored) patch has size $N_\mathrm{b} = 3\times4\times4 = 48$. In unsupervised learning, we precondition visual and textual patches with mean removal and whitening before sparse coding. We have used a dictionary size $K \approx 4N$ for both unimodal and multimodal settings. For max pooling, we use pooling factors $M$ in 10s.

In Table~\ref{tab:twitter}, we present the sentiment classification performances using linear 1-vs-all SVM. We compare performances of sparse coding methods discussed in Section \ref{sec:mult_sec}. On Twitter sentiment classification task, image and text features alone are equally useful. Thus, the image features complement the text features in unimodal feature union achieving $66.1\%$. Shared representation learned from unsupervised learning stage helps increase classification performance by 4--6\% in cross-modal features compared to unimodal features. The best feature in shallow model is multimodal feature union (Table~\ref{tab:twitter}g). Finally, we further improve the performance by building deep representations that model the correlation across the learned shallow representations. After having tried both architectures in Figures~\ref{fig:deep}a and b, we obtained the result summarized the better of two in Table~\ref{tab:twitter}. 

Table~\ref{tab:res1} presents a summary that compares the sentiment classification performances of our approach and the previous work. The authors of the PhotoTweet dataset use a combination of SentiStrength \citep{thelwall_2010} for textual features and SentiBank \citep{borth_2013} for mid-level visual features. The combined method simply concatenates the features from SentiStrength and SentiBank and does not learn shared association between modalities. We notice that our multimodal feature union (Table~\ref{tab:twitter}g) outperforms SentiStrength+SentiBank, emphasizing the importance of shared learning across multiple modalities. \cite{baecchi} use an extension of Continuous bag-of-words model  for text and denoising autoencoder for images. Again, the textual and visual features are concatenated. We compare this method with hierarchical learning with our deep multimodal sparse coding (Table~\ref{tab:twitter}h) and show that our method yields better classification result.

\begin{table}[t]
\footnotesize
\centering
\caption{Classification performance for image-text classification on PhotoTweet dataset.}
\label{tab:twitter}
\begin{tabular}{l|c}
\hline
Feature Representation                                                         & Accuracy \\ \hline
\begin{tabular}[c]{@{}l@{}}(a) Sparse coding of images\\ (Figure~\ref{fig:uni}a)\end{tabular}  & 60.91\%      \\
\begin{tabular}[c]{@{}l@{}}(b) Sparse coding of text\\ (Figure~\ref{fig:uni}b)\end{tabular}    & 58.07\%      \\
\begin{tabular}[c]{@{}l@{}}(c) Unimodal feature union\\ (Figure~\ref{fig:uni}c)\end{tabular}   & \textbf{66.10\%}      \\ \hline\hline
\begin{tabular}[c]{@{}l@{}}(d) Joint sparse coding\\ (Figure~\ref{fig:multi}a)\end{tabular}      & 70.29\%      \\
\begin{tabular}[c]{@{}l@{}}(e) Cross-modal by images\\ (Figure~\ref{fig:multi}b)\end{tabular}    & 66.64\%      \\
\begin{tabular}[c]{@{}l@{}}(f) Cross-modal by text\\ (Figure~\ref{fig:multi}c)\end{tabular}      & 62.01\%     \\
\begin{tabular}[c]{@{}l@{}}(g) Multimodal feature union\\ (Figure~\ref{fig:multi}d)\end{tabular} & \textbf{71.95\%}      \\ \hline\hline
\begin{tabular}[c]{@{}l@{}}(h) Deep multimodal sparse coding\end{tabular}    & \textbf{75.16\%}      \\\hline
\end{tabular}
\end{table}

\begin{table}[t]
\footnotesize
\centering
\caption{Classification performance comparison between multimodal sparse coding and other existing work on PhotoTweet dataset.}
\begin{tabular}{l|c|c}
\hline
\multirow{3}{*}{Feature Representation} & \multicolumn{2}{c}{Accuracy} \\ \cline{2-3} 
 & Linear & Logistic \\
 & SVM & Regr. \\ \hline
SentiStrength +  SentiBank \citep{borth_2013} & 68\% & 72\% \\
Shallow multimodal sparse coding & 71.95\% & 74.65\% \\
CBOW-DA-LR \citep{baecchi} & N/A & 79\% \\
Deep multimodal sparse coding & 75.16\% & 80.70\% \\ \hline
\end{tabular}
\label{tab:res1}
\end{table}

\section{Conclusion}

We have presented multimodal sparse coding algorithms that model the semantic correlation between modalities. We have shown that multimodal features significantly outperform the unimodal features. Our experimental results also indicate that the multimodal features learned by our algorithms is more discriminative than the feature formed by concatenating multiple unimodal features. In addition, cross-modal features computed using only one modality also lead to better performance than unimodal features. This suggests we can learn better features for one modality from joint learning of multiple modalities. The effectiveness of our approach is demonstrated in various multimedia applications such as image denoising, MED, category and sentiment classification.

{
\footnotesize
\bibliography{paper}
\bibliographystyle{iclr2016_conference}
}
\end{document}